\documentclass{article}

    \PassOptionsToPackage{numbers, compress}{natbib}

    \usepackage[final]{neurips_2021}

\usepackage[utf8]{inputenc} 
\usepackage[T1]{fontenc}    
\usepackage{hyperref}       
\usepackage{url}            
\usepackage{booktabs}       
\usepackage{amsfonts}       
\usepackage{nicefrac}       
\usepackage{microtype}      
\usepackage{xcolor}         

\usepackage[pdftex]{graphicx}

\title{
Fantastic Data and  How to Query Them\thanks{NeurIPS Data-Centric AI Workshop}}

\author{%
  Trung-Kien Tran
    \\
  Bosch Center for Artificial Intelligence,\\ Renningen, Germany\\
   \And
   Anh Le-Tuan, Manh Nguyen-Duc, Jicheng Yuan and Danh Le-Phuoc\\
  Open Distributed Systems, 
  \\Technical University of Berlin \\
}

\newcommand{\tuple}[1]{\langle#1\rangle}
\newcommand{\mi}[1]{\mathit{#1}}
 
\newcommand{\lkg}{\ensuremath{\mathcal{L}^{KG}}}

\begin{document}

\maketitle
\begin{abstract}
    It is commonly acknowledged that the availability of the huge amount of (training) data is one of the most important factors for many recent advances in Artificial Intelligence (AI). However, datasets are often designed for specific tasks in narrow AI sub areas and there is no unified way to manage and access them. This not only creates unnecessary overheads when training or deploying Machine Learning models but also limits the understanding of the data, which is very important for data-centric AI. In this paper, we present our vision about a unified framework for different datasets so that they can be integrated and queried easily, e.g., using standard query languages. We demonstrate this in our ongoing work to create a framework for datasets in Computer Vision and show its advantages in different scenarios. Our demonstration is available at \url{https://vision.semkg.org}.
\end{abstract}
\section{Vision}
\paragraph{Motivation}
The field of Artificial Intelligence (AI) advanced significantly in recent years and most of the breakthroughs are data-intensive AI or its combinations with other techniques.
Therefore, it is reasonable to argue 
that to make AI better, we need to have better ways to create (training) data and make it available to be consumed by AI models. In different AI areas, e.g., Computer Vision (CV) and Natural Language Processing (NLP) etc., and even within the same area, datasets are organised differently, e.g. stored in different formats and using different labels. This hinders not only the advancements in individual AI area but also the 
path towards Artificial General Intelligence (AGI). Let us make an analogy with the developments of human intelligence (HI) to demonstrate the importance of sharing data and knowledge. To advance in sciences and technologies, over thousands of years, we have accessed all kinds of knowledge and data, and this is because we have (written) languages. Based on previous discoveries, new ones were discovered and recorded in shareable forms. 
Similarly, it is commonly acknowledged that the current modern AI era was triggered by the availability of huge amount of data. For example, ImageNet~\cite{deng2009imagenet} helps to showcase the success of Deep Learning in CV, many datasets created from Wikipedia and other websites provide input for NLP models; although language models like GPT~\cite{radford2018improving} or BERT~\cite{devlin2018bert} do not need labelled data, it is still required in many downstream tasks. 
These datasets are so fragmented that even there exist algorithms/architectures for AGI models, there is no data to train them. This calls for the need of a unified framework to share and access datasets.
\paragraph{General Ideas}
We propose a unified framework for datasets in data-centric AI. Within this framework, not only different datasets in one AI  area but also those in  different AI areas are integrated and linked together. Existing resources such as ConceptNet~\cite{speer2017conceptnet}, Wikidata~\cite{vrandevcic2014wikidata}, have a similar goal in a sense that they integrate data from different sources and make them shareable, but they rather focus on some specific application domains, and particularly, none of them is connected with datasets in other areas such as CV. On the other hand, in CV areas, so-called \emph{scene graphs} were introduced to model the relationship between detected objects in images~\cite{johnson2015image}. However, they lack the cross-domain interoperability and cannot be queried via standard query languages. Therefore, we argue that these existing resources should be unified and new-coming resources should easily be integrated. We believe that it is very beneficial, for example, it can help to avoid distribution shift\footnote{\url{http://ai.stanford.edu/blog/wilds/}}, create more robust models for training and testing~\cite{koh2021wilds}.  Also, the shift from model tweaking to deep understanding of data furthermore requires that datasets need to be better organised~\cite{DBLP:journals/corr/abs-1909-05372}. Additionally, we believe that the term ``data'' should be extended to contain not only training data but also abstract knowledge, e.g., commonsense or causal relations~\cite{SapBABLRRSC19}.

In the next sections, we present our ongoing work to demonstrate a unified framework for datasets in CV in which images annotations are described in a knowledge graph (KG) and labels are linked to the well-known knowledge base Wikidata so that we can interlink the annotation labels across label spaces in different datasets under shared semantics. We then describe our future work and concrete steps to realise our vision.







\section{A Case Study of Vision Knowledge Graph}


\paragraph{The Vision Knowledge Graph}
To realise the ideas outlined above, we started to build a unified knowledge graph, namely VisionKG~\cite{Anh:2021}, for CV datasets (e.g, COCO~\cite{Lin:2014}, KITTI~\cite{Geiger:2013} or Visual Genome~\cite{krishna2017visual}). VisionKG is a Resource Description Framework (RDF) based knowledge graph~\citep{Hogan:2021} that contains RDF statements~\citep{Manola:2004} describing the metadata of the images and the semantic of their annotations. RDF is a standardised data model recommended by the W3C particularly for semantic data integration and as a formal representation for shared human-machine understanding. Therefore, RDF can be used to represent many semantic structures of popular label taxonomies such as Wordnet~\cite{fellbaum2010wordnet}, ConceptNet~\cite{speer2017conceptnet}, and Freebase~\cite{bollacker2007freebase} which are used in many CV datasets, e.g., Imagenet~\cite{Krizhevsky:2012}, OpenImage~\cite{OpenImages} and VisualGenome~\cite{krishna2017visual}.

\begin{figure}[ht!]
    \centering
    \includegraphics[width=1.0\textwidth]{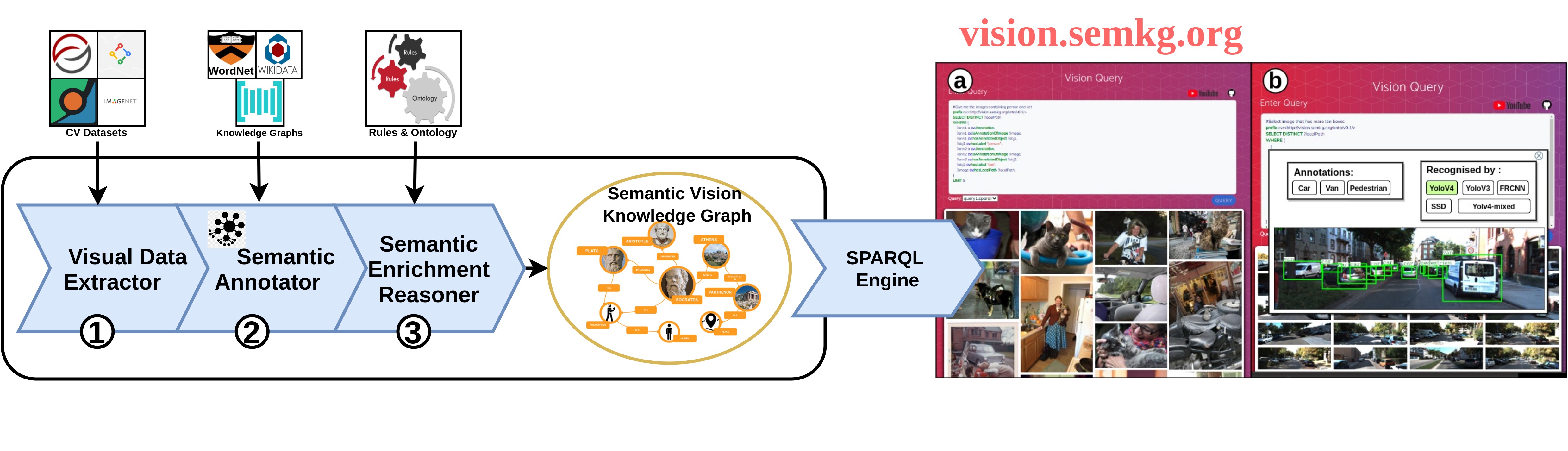}
    \caption{The overview of VisionKG}
    \label{fig:vision_kg}
\end{figure}

\begin{figure}[ht!]
    \centering
    \includegraphics[width=\textwidth]{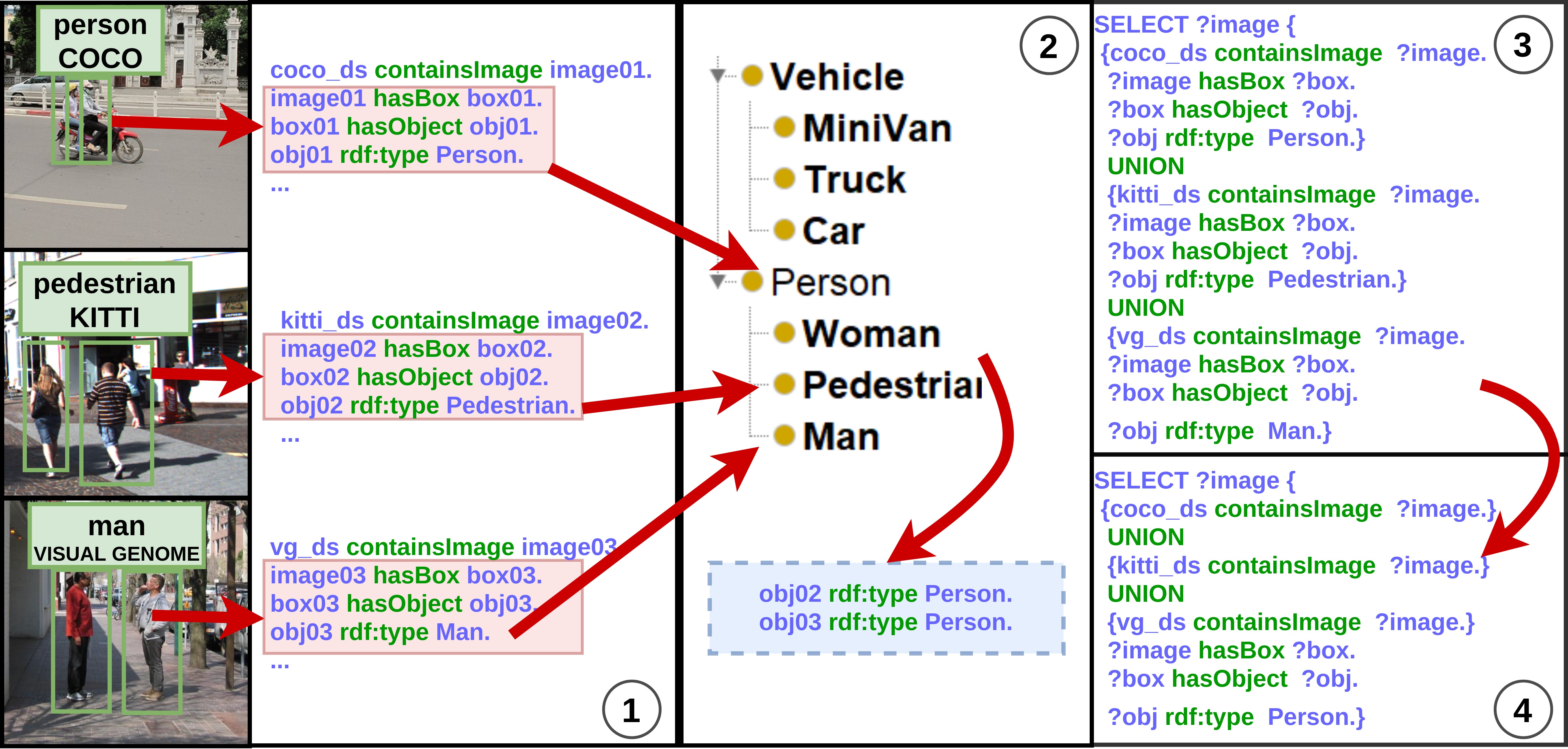}
    \caption{Mapping labels in COCO, KITTI, and Visual Genome to classes in the knowledge base (1). Expanding the labels according to the class hierarchy (2). And examples of two equivalent queries to obtain images that contain 
    {Person} from COCO, KITTI, and Visual Genome datasets (3,4). 
    }
    \label{fig:example}
\end{figure}

Figure~\ref{fig:vision_kg} illustrates the process of creating VisionKG. 
We first collect CV datasets and extract their annotation labels~\textcircled{\raisebox{-0.9pt}{1}}.
To create a unified data model for the annotation labels and visual features, we follow the Linked Data principles~\citep{Bizer:2011} and use the RDF data model.
The data entities (e.g, images, boxes, labels) are named using Uniform Resource Identifiers (URI). 
RDF data model allows the data to be expressed using   \emph{triples} of the form $\mi{\tuple{subject,predicate, object}}$.
For example, to describe ``an image contains a bounding box for a person" in COCO dataset, we first assign unique URIs, e.g., \textit{vision.semkg.org/img01} and \textit{vision.semkg.org/box01}, for the image and the bounding box, respectively to create the following triples for such image:
$\mi{\tuple{img01, hasBox, box01}, \tuple{box01, hasObject, obj01}, \tuple{obj01,rdf:type,Person}}$.  The predicates \textit{hasBox}, \textit{hasObject}, and \textit{rdf:type} are predefined, in which \textit{rdf:type} is used to express an object/image belongs to a specific class/type, e.g., \textit{Person}, in the knowledge base; for simplicity, we skip the prefix in these triples. 
%
%
Furthermore, we add metadata and semantic annotations for the images, e.g., where the images come from or what are the relations of the boxes in an images (Figure~\ref{fig:vision_kg}~\textcircled{\raisebox{-1pt}{2}}). The earlier provides the semantic relationships to facilitate semantic reasoning capability in below.
\paragraph{Unifying and Expanding Labels}
Since labels, attributes, and relationships in COCO, KITTI, and Visual Genome datasets are either just text or mapped to WordNet~\cite{fellbaum2010wordnet}, we link them to the corresponding predicates and classes in Wikidata.\footnote{\url{https://www.wikidata.org/}} Wikidata is an open knowledge graph commonly used in other application domains, and therefore integrating datasets via Wikidata make them available also to other domains. Another advantage is that we can utilise the existing class hierarchy as shown in Figure~\ref{fig:example} to add more labels to existing datasets using a semantic \emph{reasoner} to expand/materialise the labels. For example, a box that is labelled as \textit{a pedestrian} is also annotated as \textit{a person}  (Figure~\ref{fig:example}~\textcircled{\raisebox{-1pt}{1}}) because in the hierarchy of the knowledge base, \textit{pedestrian} is a subclass of \textit{person} (Figure~\ref{fig:example}~\textcircled{\raisebox{-1pt}{2}}). Hence, our knowledge graph can interlink the annotation labels across label spaces under shared semantic understanding.  Along with the semantic relationships, thanks to the URI representation of image instances, different types of annotations for different learning tasks can be linked together via the URI of the images. For example, COCO~\cite{Lin:2014}, COCO-Stuff~\cite{caesar2018coco}, and VisualGenome~\cite{krishna2017visual} share many common images for different annotations in object detection, semantic segmentation, and visual relationship detection.

\paragraph{Use Cases}
VisionKG enables a more effective way to organise training data and offers more robust ways to analyse and evaluate trained Deep Neural Networks (DNNs). Specifically, datasets and analysis can be done using rich semantic query languages such as SPARQL.\footnote{\url{https://www.w3.org/TR/rdf-sparql-query/}}
The SPARQL query language provides users the ability to describe queries using 
RDF statements which are similar to those in the SQL language. The first use case is that VisionKG can be used to obtained mixed-datasets in an elegant way. For instance, one can query for images of \textit{person} from COCO, KITTI, and Visual Genome using a simple query (Figure~\ref{fig:example}~\textcircled{\raisebox{-1pt}{4}}) instead of the more complex query (Figure~\ref{fig:example}~\textcircled{\raisebox{-1pt}{3}}) that covers all possible cases: \textit{pedestrian} in KITTI or \textit{man} in Visual Genome. This is possible because we already aligned labels of existing datasets with the common taxonomy and expanded them according to the taxonomy hierarchy.
 Similarly, one can conveniently create mixed-test data. In advanced settings, users can have more fine-grained criteria for retrieving images such as the query for \emph{"images that contain a person holding a cat"} shown in Figure 1~\textcircled{\raisebox{-0.9pt}{a}}. Other use cases include analytic queries to have deeper understanding about the dataset and performances of DNNs on some particular sub-set of the data.
Figure 1~\textcircled{\raisebox{-0.9pt}{b}} illustrates a complex query to \emph{"search for the trained models to detect \textit{Car} on images of \textit{Car} in crowded traffic scenes or in mountain areas"}. More example SPARQL queries can be found in our online VisionKG system.\footnote{ \url{https://vision.semkg.org}}






\section{Towards a Unified Framework for Querying Data on The Fly}
Our case  study shows 
how the data can be organised in a unified way for both training and test phases. This approach advocates for an AI strategy that shifts from emphasis on proprietary datasets, to the sharing of data across entities for knowledge creation, namely building a knowledge graph guiding learning algorithms, called \textbf{Learning Knowledge Graph} \lkg.  Such a strategy leads us to look towards building a unified framework to facilitate the ability to query data in a declarative fashion for data loading, training, validation and test phases of Machine Learning pipelines. These pipelines are not restricted to neural networks but also learning approaches such as  probabilistic logic programming~\cite{Manhaeve:2021} or a fusion of various learning algorithms, e.g.~\cite{Le-Phuoc:2021}. Such a framework with \lkg will enable a learning algorithm to dynamically retrieve data programmatically. Hence, it does not have to assume training/validation/test sets to be fixed. This might lead to more robust models for a set of learning architectures. In particular, \lkg can give such a training algorithm the access to a much bigger set of training samples than those in a traditional training pipeline.  For example, we are planning to integrate into \lkg a larger set of classification samples than those in the currently popular datasets such as 14 millions images of ImageNet~\cite{deng2009imagenet} or even 300 millions images of JFT-300M~\cite{sun2017revisiting} with probably open label spaces, e.g.~\cite{Xiangyun:2020,koh2021wilds,Rui:2021}. Note that \lkg will also consider unsupervised, self-supervised and weakly-supervising learning algorithms, hence, bigger sets of images such as  Youtube-8M~\cite{Abu:2016}, YFCC100M~\cite{Thomee:2016} and LAION-400-MILLION~\footnote{https://laion.ai/laion-400-open-dataset/} are also to be integrated. In this light, it will be much more challenging to build a learning architecture as it has to deal with ever-growing training/validation/test sets.

Towards this vision, instead of focusing on learning architectures, i.e., writing new models instead of understanding the nature of the learning tasks and its semantic relationships to the potential data being learnt, 
we focus on how to represent semantic relations, "know-how", context and domain knowledge, then make them queryable so that the training algorithms can exploit such a human-machine understandable knowledge to  build the desired models by just specifying the expected outputs in a declarative fashion. For instance,  to build a model for scene understanding, $\mathcal{L^{KG}}$ can specify  how an image scene can be represented as a scene graph constructing from edges (e.g., RDF statements) representing relationships among objects which can be learnt via the models to detect visual relationships, e.g~\cite{lu:2016}, called VRD models.  A training algorithm to build such VRD models will have to rely on datasets such as Visual Genome~\cite{krishna2017visual} or VrR-VG\cite{Yuanzhi:2019} which have various semantic relationships with datasets for training object detection/segmentation and classifications such as COCO and ImageNet. Interestingly,  VisionKG~\cite{Anh:2021} illustrated that such relationships can also be connected to natural language data sources such as Wordnet~\cite{miller1995wordnet} and Conceptnet~\cite{speer2017conceptnet}, Wikidata~\cite{vrandevcic2014wikidata}, and Freebase~\cite{bollacker2007freebase}. Along the same line, various datasets can help to facilitate knowledge embedding associated with natural language ones such as CLIP~\cite{Alec:2021}, VisualComet~\cite{park2020visualcomet} and VCR~\cite{Rowan:2018}. On the other hand, there are many text-based datasets that can be enriched with visual data such as~\cite{Srinivasan:2021},~\cite{Gonz:2021} and~\cite{Yongfei:2021}. To this end, the next challenge for our framework is how to leverage such rich correlated information among datasets and learning tasks to automate the training algorithms to make it faster, more efficient and more robust in building AI component powered by \lkg.  


Our such knowledge-centric approach hints that companies and organizations should focus on preparing knowledge-driven AI development pipeline. This shift  shapes organizations’ HR strategy on how to build an AI team that embrace the potential capabilities of \lkg . 
While some companies will still require hiring large cohorts of rare and expensive data engineers and scientists, knowledge engineers can offer alternative resources to scale up AI development pipelines. The joint force among them can leverage the knowledge discovery in such pipelines to automate various phases of the development. For instance, the work such as Taskonomy~\cite{Amir:2018} and Taskology~\cite{Yao:2021} can facilitate the processes of building taxonomies/ontologies of the learning tasks so that the relationships among them can be exploited to build more efficient multi-task learning pipelines. Moreover, our knowledge-driven approach will lean towards a similar automation approach of the Software 2.0~\cite{Re:2018}, DynaBench~\cite{Douwe:2021}, DynaBoard~\cite{Zhiyi:2021} whereby the knowledge engineers and domain experts can specify application specifications, tests and benchmarks along with domain knowledge as the input to automate the AI development phases, such as data cleaning, data integration, label creation and model search, model selection and model testing.

\section*{Acknowledgements}
This work was funded by the German Research Foundation (DFG) under the COSMO project (ref. 453130567), the German Ministry for Education and Research via The Berlin Institute for the Foundations of Learning and Data (BIFOLD, ref. 01IS18025A and ref. 01IS18037A), and the German Academic Exchange Service (DAAD, ref. 57440921). We thank Jannik Str\"otgen for his valuable feedback on this work.

\bibliographystyle{ACM-Reference-Format}
\bibliography{main}


\end{document}